\documentclass{article}

\usepackage{arxiv}

\usepackage[usenames,dvipsnames]{xcolor}
\usepackage[utf8]{inputenc} % allow utf-8 input
\usepackage[T1]{fontenc}    % use 8-bit T1 fonts
\usepackage{hyperref}       % hyperlinks
\usepackage{url}            % simple URL typesetting
\usepackage{booktabs}       % professional-quality tables
\usepackage{amsfonts}       % blackboard math symbols
\usepackage{nicefrac}       % compact symbols for 1/2, etc.
\usepackage{microtype}      % microtypography
\usepackage[inline]{enumitem}
\usepackage{lipsum}
\usepackage{xspace}
\usepackage{tikz}
\usepackage{mwe}

\usepackage{mymacros}

\hypersetup{
    colorlinks,
    citecolor=Violet,
    linkcolor=Red,
    urlcolor=Blue}

\usetikzlibrary{backgrounds}
\usetikzlibrary{decorations}
\usetikzlibrary{decorations.pathreplacing}
\usetikzlibrary{fit}
\usetikzlibrary{positioning}

\title{The Quo Vadis submission at Traffic4cast 2019}

\author{
  Dan Oneață%
  \thanks{D. Oneață, M. Stănescu and H. Cucu are with University Politehnica of Bucharest, Romania. E-mail: \texttt{dan.oneata@gmail.com}.}
  \And
  Cosmin George Alexandru
  \And
  Marius Stănescu
  \AND
  Octavian Pascu
  \And
  Alexandru Magan
  \And
  Adrian Postelnicu
  \And
  Horia Cucu
}

\date{}

\renewcommand{\headeright}{Extended abstract}
\renewcommand{\undertitle}{Extended abstract}

\newcommand{\bias}{\mathsf{B}}
\newcommand{\bloc}{\mathsf{L}}
\newcommand{\bweek}{\mathsf{W}}
\newcommand{\bhour}{\mathsf{H}}
\newcommand{\bmonth}{\mathsf{M}}
\newcommand{\tempreg}{\mathsf{TR}}
\newcommand{\data}{\mathbf{D}}

\newcommand{\mnaive}{\textsf{\small na\"ive} \xspace}
\newcommand{\mzeros}{\textsf{\small zeros} \xspace}
\newcommand{\maverage}{\textsf{\small average} \xspace}
\newcommand{\mseasonal}{\textsf{\small seasonal} \xspace}

\newcommand{\nero}{\texttt{\color{Plum}Nero}\xspace}
\newcommand{\codemodel}[1]{\texttt{\color{Plum}#1}\xspace}

\begin{document}
\maketitle

\begin{abstract}
    We describe the submission of the Quo Vadis team to the Traffic4cast competition,
    which was organized as part of the NeurIPS 2019 series of challenges.
    Our system consists of a temporal regression module, implemented as 1$\times$1 2d convolutions, augmented with spatio-temporal biases.
    We have found that using biases is a straightforward and efficient way to include seasonal patterns and to improve  the performance of the temporal regression model.
    Our implementation obtains a mean squared error of $9.47\times 10^{-3}$ on the test data, placing us on the eight place team-wise.
    We also present our attempts at incorporating spatial correlations into the model;
    however, contrary to our expectations, adding this type of auxiliary information did not benefit the main system.
    Our code is available at \url{https://github.com/danoneata/traffic4cast}.
\end{abstract}

% keywords can be removed

\section{Introduction}
\label{sec:introduction}

This paper describes our entry in the Traffic4cast challenge.
The goal of the competition was to forecast traffic activity---heading, speed, volume---for the next 5, 10 and 15 minutes.
The data is represented as a set of image frames, with each pixel encoding the average information obtained from a 100 m$^2$ area over a 5-minute time interval.
The measurements are collected from three large cities (Berlin, Moscow, Istanbul) over the span of one year.

The challenge is unique compared to most prior work,
the two closest related tasks being
\ia traffic forecasting \cite{cai_spatiotemporal_2016,cheng_deeptransport:_2018,li_diffusion_2018} and
\ib video prediction \cite{xue_visual_2016,zhou_view_2016,vondrick_generating_2017}.
The traffic forecasting task, in its usual formulation, is concerned with predicting traffic activity at a small number of locations in a city (typically hundreds of positions),
whereas in the actual challenge we are operating on a densely sampled grid (495 $\times$ 436 locations);
an exception is the work of Yu \etal \cite{yu_spatiotemporal_2017} which also represent the data as images.
Compared to the video prediction task, the current task differs in that
the data has fine details (streets and roads),
the prediction is location dependent and heavily influenced by this underlying structure, and finally
motion information is encoded in the data (through the heading and speed channels).

The manuscript is structured as follows.
In section \ref{sec:main-system-description} we describe our best system and in section \ref{sec:results} we report the experimental results.
In section \ref{sec:alternative-attempts} we present alternative systems, which were outperformed by our main system,
but which we nevertheless found informative. Furthermore,
in the same section, we briefly mention other interesting findings from our experiments.
Finally, section \ref{sec:conclusions} concludes the paper.

\section{Main system}
\label{sec:main-system-description}

In this section we describe our submitted system, which consists of two parts:
a temporal regression module and biases.

\textbf{Temporal regression ($\tempreg$) module.}
The backbone of our system is a model which performs temporal regression---%
it predicts the future values at a location $(x, y)$ by looking at the historical values at the same location $(x, y$).
This module is implemented as a composition of 1$\times$1 2d convolutions interspersed with non-linear activation functions.
We have experimented with four types of activation functions (ReLU, ELU, SELU, leaky ReLU)
and have settled on ELU activations \cite{clevert_fast_2016} for the final submission.

\textbf{Biases ($\bias$)}.
To capture location-specific and temporal patterns,
we augment the model with spatial and temporal biases.
The biases are values that do not depend on the historical values,
but they are based on other information,
such as the coordinates of the pixel or the (absolute) time when the prediction is made (\eg, hour, day of week, month).
The implementation uses learnable parameters represented as matrices.
The types of biases we have considered are described below:
\begin{center}
    \setlength{\tabcolsep}{15pt}
    \begin{tabular}{lllr}
        \toprule
        Domain & Resolution       & Notation & Size \\
        \midrule
        space / location & pixel       & $\bias_{\bloc}$  & $495 \times 436$ \\
        time  & hour        & $\bias_{\bhour}$ & $12$ \\
        time  & day of week & $\bias_\bweek$   & $7$  \\
        time  & month       & $\bias_\bmonth$  & $12$ \\
        \bottomrule
    \end{tabular}
\end{center}

Several biases can be additively combined. However, such combinations omit interactions between the domains.
For example,
the traffic on Sunday is generally lower than on Monday,
but at a location on the outskirts it may remain constant as people are returning to the city.
In order to capture these dependencies,
we also use biases over combinations of the four domains.
We denote those using the operator $\times$, to reflect the size of the parameter matrix.
% $\bias_{\bloc\times\bhour}$ or $\bias_{\bloc\times\bweek}$.
As we discuss in section \ref{sec:results}, we experimented with multiple combinations,
but for the final submission we have used the following three types of biases:
$\bias_{\bloc\times\bhour}$,
$\bias_{\bweek\times\bhour}$ and
$\bias_{\bmonth}$.

To summarize, our main system sums the prediction of the temporal regression module, $\tempreg$, with the three biases:
\[
    y(x, y, t: t + 3)
        = \tempreg(\data[x, y, t - h: t])
        + \bias_{\bloc\times\bhour}[x, y, t_\bhour]
        + \bias_{\bweek\times\bhour}[t_\bweek, t_\bhour]
        + \bias_{\bmonth}[t_\bmonth],
\]
where
$\data$ denotes the data,
$h$ is the history length,
$(x, y)$ is the location of the prediction,
$t$ denotes the time, with $t_H$ being the hour, $t_W$ the week and $t_M$ the month.

The model optimizes the mean squared error (MSE) loss---the same metric used for evaluation by the competition---using the Adam algorithm.
We perform early stopping on the validation set and learning rate scheduling:
the learning rate is decreased by a factor of 5 if the performance stops improving for more than two epochs.
Each epoch uses 20\% randomly sampled days from the training dataset.
Given that only a part of the time slices in a day are used at test time (five time slices),
for the final model we learn to predict only on at those particular target time slots.
Training with target slices from any time was also performed, but we did not notice significant differences in results.
Our solution is implemented in Python, using PyTorch, and can be found at
\url{https://github.com/danoneata/traffic4cast}.
The main model is named \nero (see \texttt{models/nn.py}).

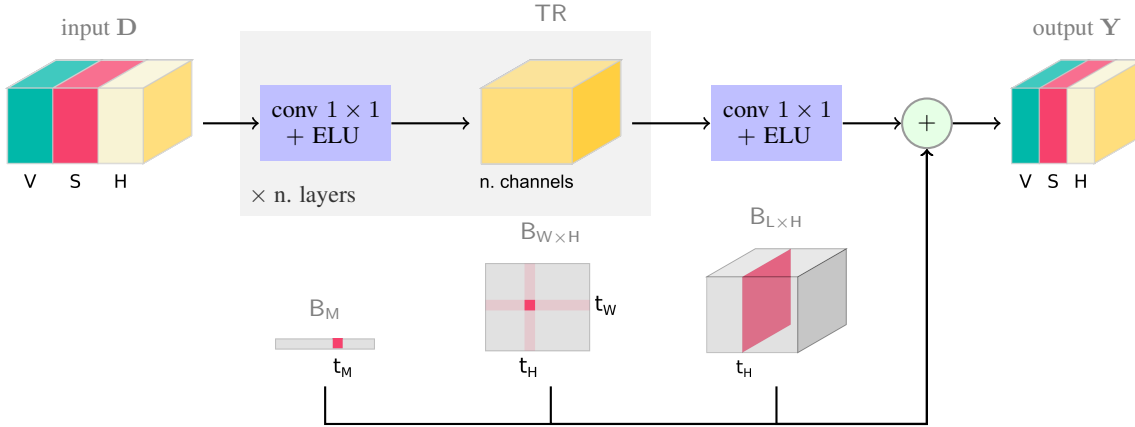
\begin{figure}
    \centering
    \begin{tikzpicture}[
        font=\footnotesize,
        auto,
        thick,
        box/.style={text width=1.5cm, minimum width=0.5cm, align=center, fill=blue!25, minimum height=1.0cm},
    ]
    \node[label={\color{gray}input $\data$}] (input) at (0, 0) {\includegraphics[height=1.7cm]{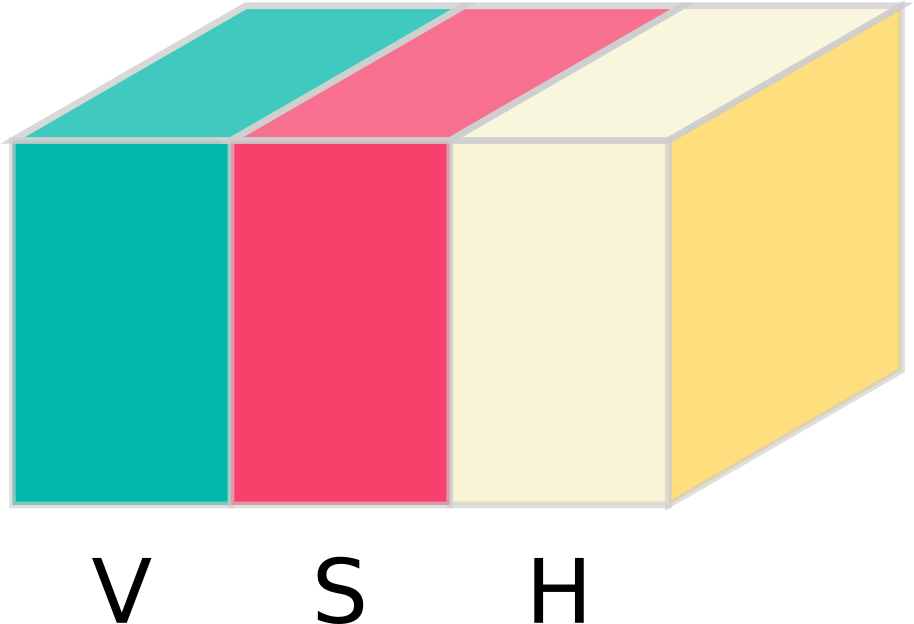}};
    \node[box] (conv1) at (3, 0) {conv $1\times1$ $+$ ELU};
    \node[label={[label distance=7pt]\color{gray}$\tempreg$}] (middle) at (6, 0) {\includegraphics[height=1.7cm]{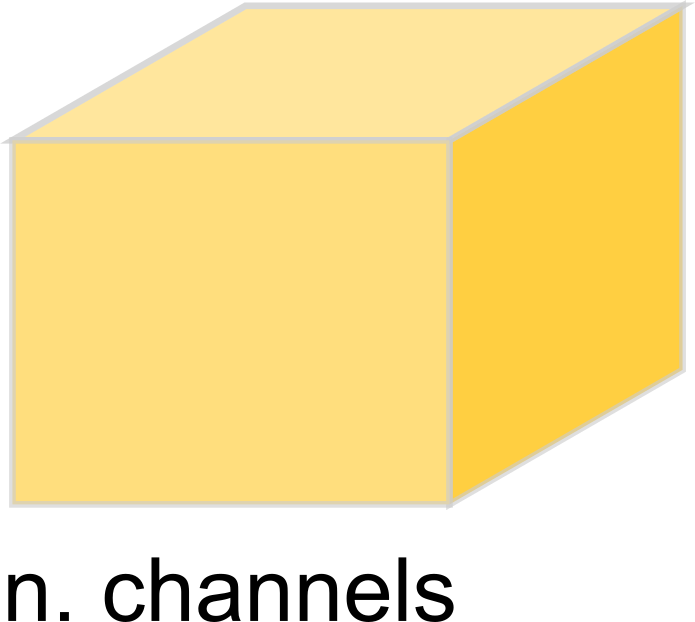}};
    \node[box] (conv2) at (9, 0) {conv $1\times1$ $+$ ELU};
    \node[circle, fill=green!10, draw=black!40] (add) at (11, 0) {$+$};
    \node[label={\color{gray}{output $\mathbf{Y}$}}] (output) at (13, 0) {\includegraphics[height=1.7cm]{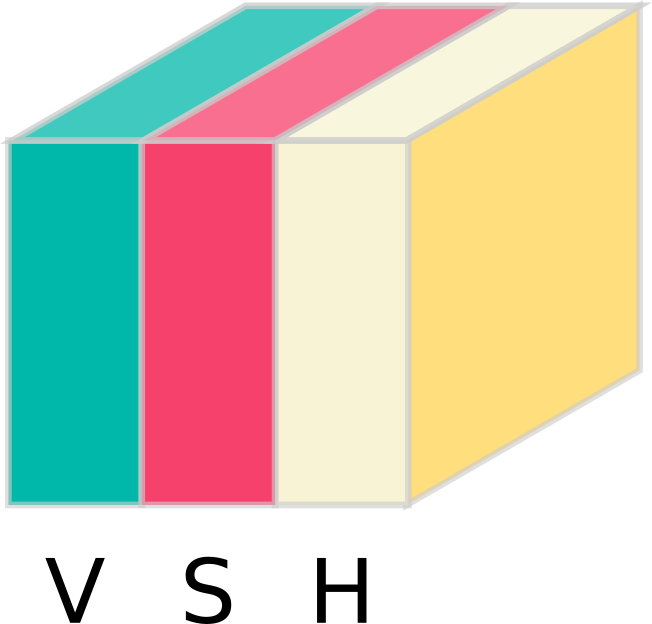}};
    \node[anchor=south, label={\color{gray}$\bias_{\bloc\times\bhour}$}]  (blxh) at (9, -3.5) {\includegraphics[height=1.7cm]{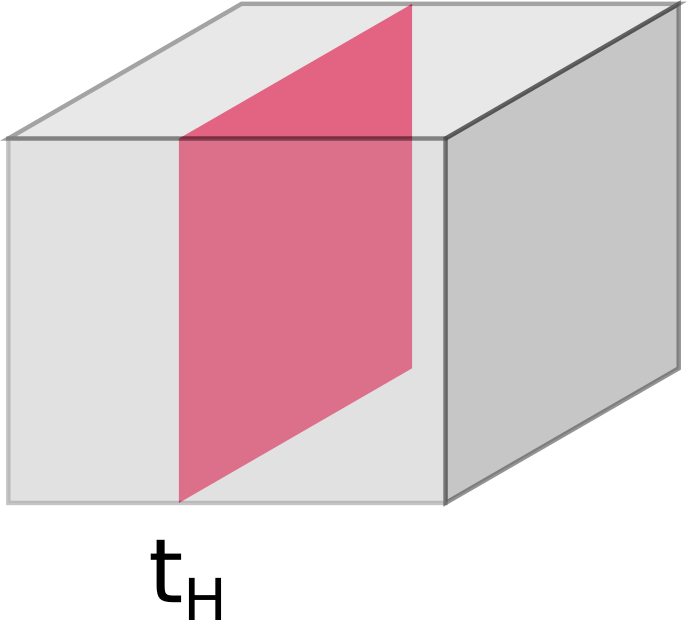}};
    \node[anchor=south, label={\color{gray}$\bias_{\bweek\times\bhour}$}] (bwxh) at (6, -3.5) {\includegraphics[height=1.5cm]{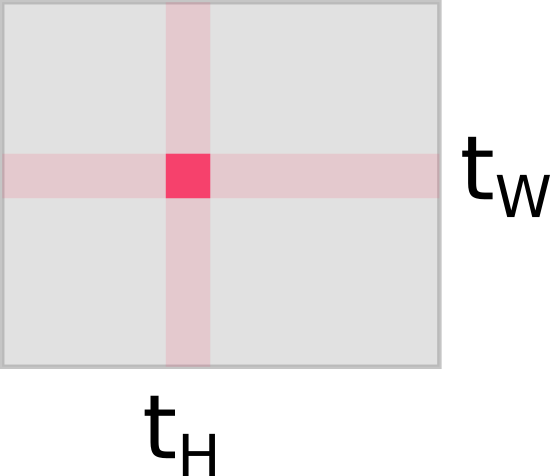}};
    \node[anchor=south, label={\color{gray}$\bias_{\bmonth}$}]            (bm)   at (3, -3.5) {\includegraphics[height=0.5cm]{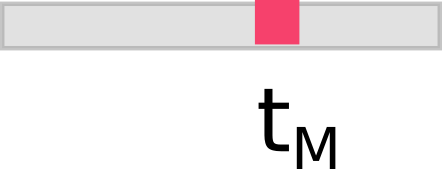}};

    \coordinate (x) at (9, -4);
    \coordinate (y) at (6, -4);
    \coordinate (z) at (3, -4);

    \draw[->] (input)  -- (conv1);
    \draw[->] (conv1)  -- (middle);
    \draw[->] (middle) -- (conv2);
    \draw[->] (conv2)  -- (add);
    \draw[->] (blxh)   -- (x) -| (add);
    \draw[->] (bwxh)   -- (y) -| (add);
    \draw[->] (bm)     -- (z) -| (add);
    \draw[->] (add)    -- (output);

    % \draw [decorate,decoration={brace,mirror,amplitude=10pt,raise=4pt}] (2, -1) -- node [below=20pt] {$\times$ n. layers} (7, -1);
    \begin{pgfonlayer}{background}
      \node[inner sep=7pt, fill=gray!10, fit=(conv1) (middle)] (plate) {};
      \node[above right, color=gray!40!black] at (plate.south west) {$\times$ n. layers};
    \end{pgfonlayer}
    \end{tikzpicture}
    \caption{%
      The architecture of our system.
      To obtain the final prediction $\mathbf{Y}$ we pass the input data $\data$ through the temporal regression network $\tempreg$ and
      add biases for location ($\bloc$) and time (hour $\bhour$, week $\bweek$, month $\bmonth$).%
    }
\end{figure}

\newpage 
\section{Results}
\label{sec:results}

In this section we present the experimental results of our system and contrast it to four baseline models:
\begin{itemize}
    \item \mzeros -- predicts only zeros, \ie, the values for frames $t+1, t+2, t+3$ are all zeros;
    \item \mnaive -- predicts the last frame, \ie, the values for frames $t+1, t+2, t+3$ is the value at $t$;
    \item \mseasonal -- averages the predictions from the previous three days at the same time;
    \item \maverage -- running average predictions using the last three three frames.
\end{itemize}

In addition to the main system (denoted by $\tempreg + \bias$),
we include a variant of the system without the biases (denoted by $\tempreg$).
The results are presented in Table \ref{tab:main-results}.
First, we observe that the proposed model significantly outperforms the baseline models.
Second, we notice that the biases clearly help for all channels and cities.

\begin{table}[b] 
    \setlength{\tabcolsep}{4pt}
    \newcommand{\ii}[1]{\footnotesize{\color{CadetBlue}#1}}
    \centering
    \caption{%
        Mean squared error ($\times 10^{-3}$) for baseline methods (rows 1--4), variants of our approach (rows 5--6), and the best entry in the challenge (row 7).
        For the validation data we report detailed results for the three cities (Berlin, Moscow, Istanbul) and three channels (volume -- V, speed -- S, heading -- H).
        For the test data we report the mean over the nine combinations (cities $\times$ channels) as computed on the leaderboard.
    }
    \begin{tabular}{ll|rrr|rrr|rrr|rrr|rr}
        % \toprule
        &            & \multicolumn{3}{c}{Berlin} & \multicolumn{3}{c}{Moscow} & \multicolumn{3}{c}{Istanbul} & Berlin & Moscow & Istanbul & valid. & test \\
        % \cmidrule(lr){2-4} % \cmidrule(lr){5-7} % \cmidrule(lr){8-10}
        & Method     & V      & S      & H      & V      & S      & H      & V      & S      & H      & mean   & mean   & mean    & mean   & mean \\
        \midrule
        % Old results from the Wiki:
        % \mzeros    & 0.0005 & 0.0151 & 0.0271 & 0.0003 & 0.0169 & 0.0621 & 0.0005 & 0.0138 & 0.0552 & 0.0143 & 0.0265 & 0.0232    & 0.0213    & 0.0216 \\
        % \mnaive    & 0.0004 & 0.0095 & 0.0300 & 0.0001 & 0.0095 & 0.0583 & 0.0001 & 0.0053 & 0.0465 & 0.0133 & 0.0227 & 0.0173 & 0.0178 \\
        \ii{1} & \mzeros            & 0.5 & 15.1 & 27.1 & 0.3 & 17.4 & 64.3 & 0.5 & 14.3 & 56.9 & 14.3 & 27.3 & 23.9 & 21.83 & 21.69 \\
        \ii{2} & \mnaive            & 0.4 & 9.5  & 29.9 & 0.1 & 9.6  & 60.2 & 0.1 & 5.3  & 48.2 & 13.3 & 23.3 & 17.9 & 18.17 & \\
        \ii{3} & \mseasonal         & 0.5 & 10.2 & 21.8 & 0.2 & 10.5 & 46.0 & 0.2 & 8.0  & 38.1 & 10.8 & 18.9 & 15.4 & 15.03 & \\
        \ii{4} & \maverage          & 2.6 & 9.5  & 20.2 & 4.7 & 10.7 & 37.5 & 3.1 & 7.6  & 31.7 & 10.8 & 17.6 & 14.1 & 14.17 & \\
        \midrule
        % python evaluate3.py -m nero-8-32-nobias --channel Volume Speed Heading -v -c Berlin   -p output/models/nero-8-32-nobias_Volume_Speed_Heading_Berlin.pth
        % python evaluate3.py -m nero-8-64-nobias --channel Volume Speed Heading -v -c Moscow   -p output/models/nero-8-64-nobias_Volume_Speed_Heading_Moscow.pth
        % python evaluate3.py -m nero-4-64-nobias --channel Volume Speed Heading -v -c Istanbul -p output/models/nero-4-64-nobias_Volume_Speed_Heading_Istanbul.pth
        \ii{5} & $\tempreg$         & 0.3 & 5.3  & 16.1 & 0.1 & 5.4  & 33.3 & 0.1 & 3.0  & 26.4 & 7.2  & 12.9 & 9.8  & 9.96  & \\
        \ii{6} & $\tempreg + \bias$ & 0.3 & 5.0  & 15.6 & 0.1 & 5.1  & 31.5 & 0.1 & 2.8  & 25.0 & 7.0  & 12.2 & 9.3  & 9.50  & 9.47 \\
        \midrule
        \ii{7} & winner             &     &      &      &     &      &      &     &      &      &      &      &      &       & 9.01 \\
        % \bottomrule
    \end{tabular}
    \label{tab:main-results}
\end{table}

\begin{figure}
    \centering
    \includegraphics[clip,trim=1.5cm 2.1cm 1.5cm 2.1cm,width=0.9\textwidth]{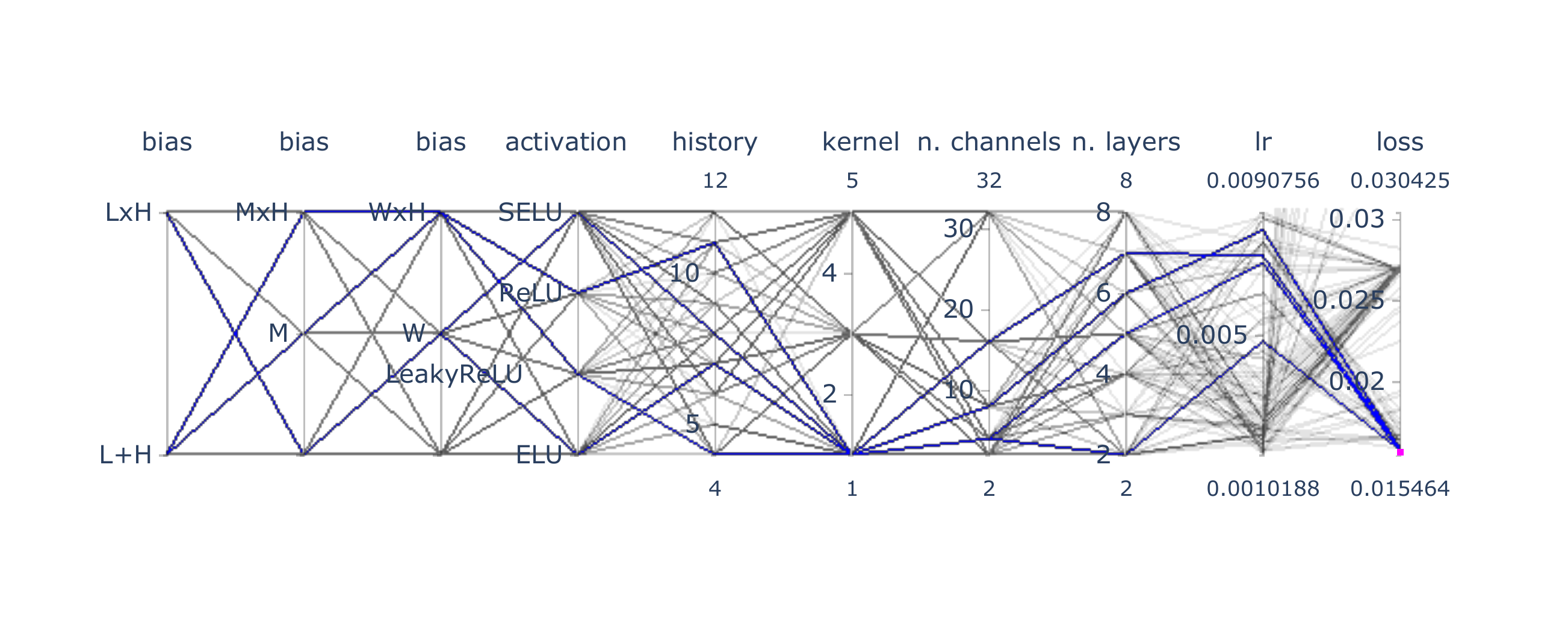} \\
    Berlin \\
    \vspace{0.5cm}
    \includegraphics[clip,trim=1.5cm 2.1cm 1.5cm 2.1cm,width=0.9\textwidth]{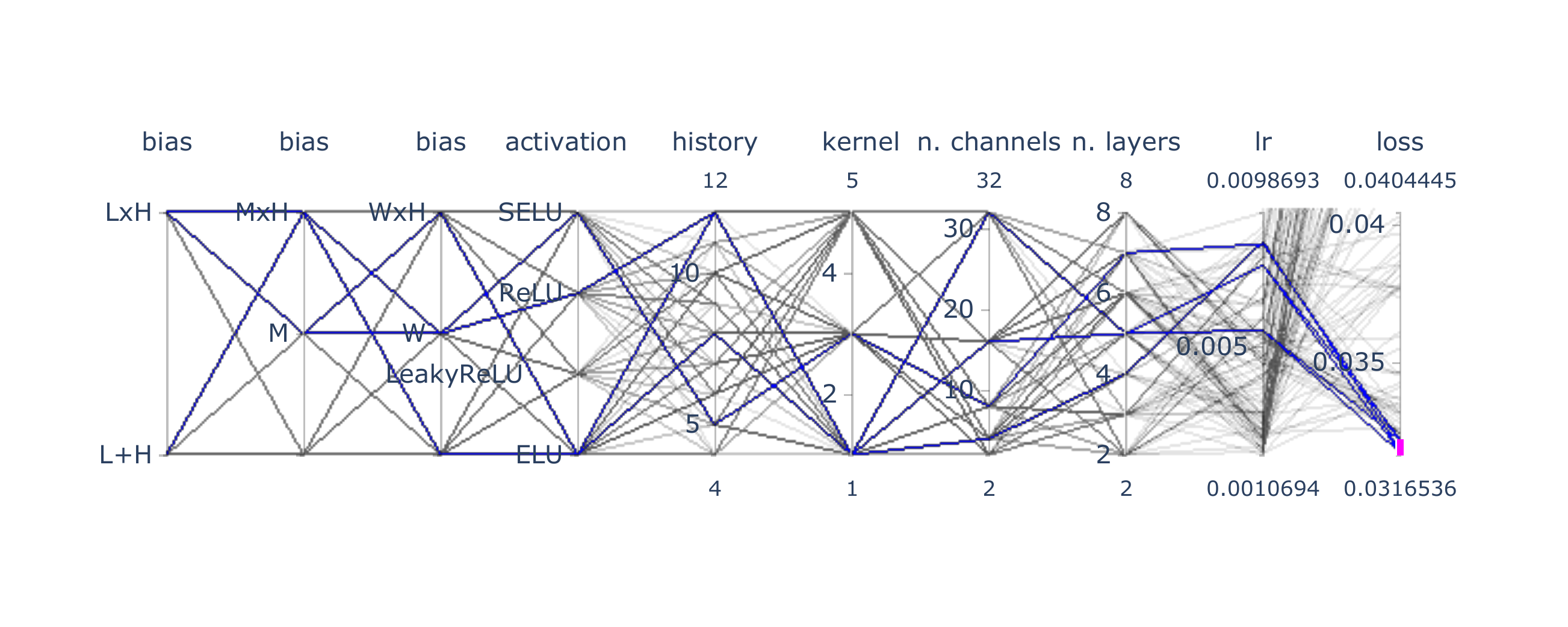} \\
    Moscow \\
    \vspace{0.5cm}
    \includegraphics[clip,trim=1.5cm 2.1cm 1.5cm 2.1cm,width=0.9\textwidth]{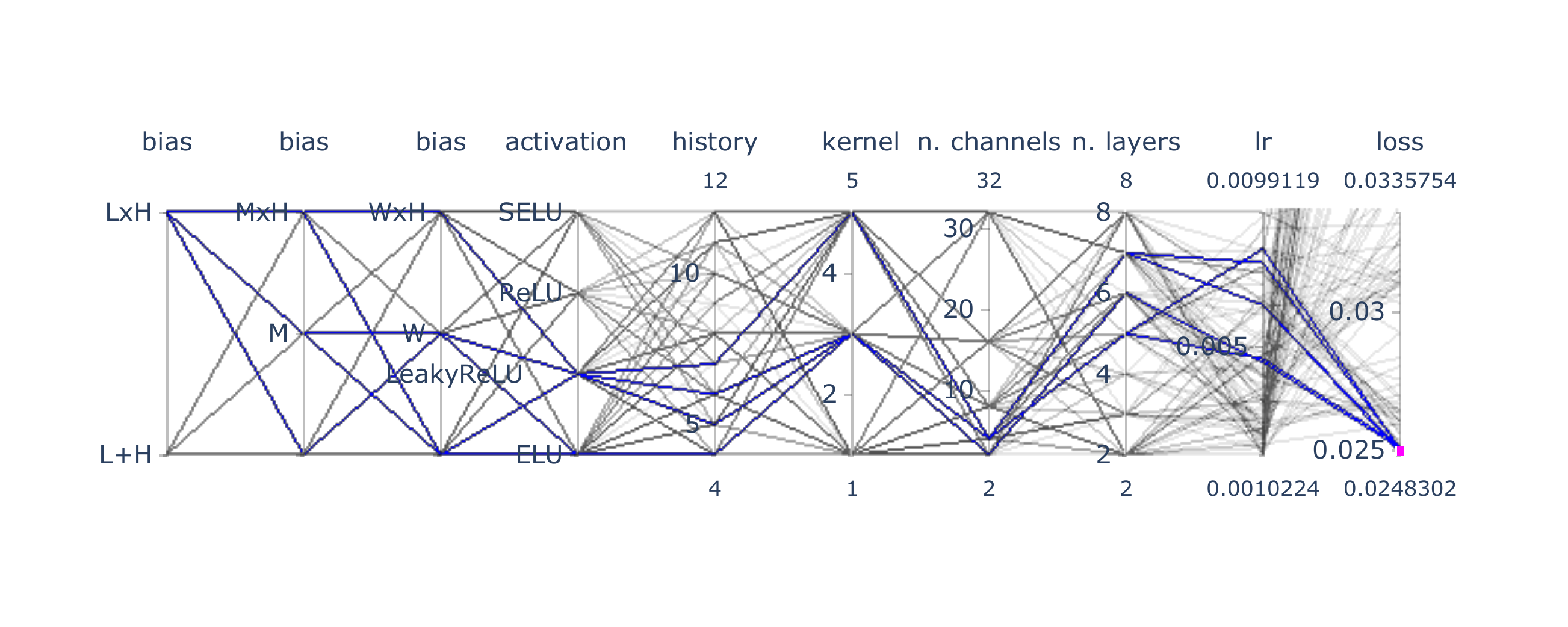} \\
    Istanbul
    \caption{
        Hyper-parameter tuning results on the validation set for the main model applied only on the heading channel for the three cities.
        Each parallel axis encodes the values of a hyper-parameter, except the last one which shows the loss (mean squared error).
        A path through the plot represents a hyper-parameter configuration and the achieved loss.
        The blue paths indicate the top five configurations.
    }
    \label{fig:hyperopt}
\end{figure}

\textbf{Exploring the hyper-parameters.}
In order to understand the sensitivity of the model to the hyper-parameters we have run systems with multiple hyper-parameter combinations.
We used the HyperBand algorithm \cite{li_hyperband:_2018} from the \texttt{HpBandSter} Python library.
This method randomly samples hyper-parameter values, but as opposed to other random search methods,
it uses different budget values---it tries many configurations on lower computational budgets,
from which only the most promising ones are explored further with larger budgets.
The hyper-parameters we have considered include the learning rate of the optimizer,
the number of convolutional layers,
number of channels of the convolutional layer,
kernel size, history and biases.
For this set of experiments, we evaluated by using only the heading channel,
as it was dominating the average error.
The results are summarized in figure \ref{fig:hyperopt}.

The results showed that the choice of the learning rate matters,
although we observed that ELU or SELU activations allow for a broader range of the learning rate.
A smaller kernel (size 1) was generally better, with the exception of Istanbul.
For the final model we used kernel size of 1 for all three cities and ELU activations. The number of channels and layers were tuned independently for each city.
% \begin{center}
%     \includegraphics[width=0.3\textwidth]{heatmap-Berlin}
%     \includegraphics[width=0.3\textwidth]{heatmap-Moscow}
%     \includegraphics[width=0.3\textwidth]{heatmap-Istanbul}
% \end{center}

\section{Alternative attempts}
\label{sec:alternative-attempts}

Besides the main system, several other alternative architectures have been tried.
The main goal was to incorporate spatial structure into the model,
but, to our surprise, none of these attempts performed better than the main system,
which is agnostic to spatial correlations.
In this section we describe two of the most interesting approaches.
More undocumented efforts can be found in the code in the \texttt{models/nn.py} and from the \texttt{master} and other branches.

\textbf{Motivation.}
We have observed that there are large displacements of the non-zero values from one frame to the other.
For example, at the outskirts of the city there are few vehicles, which consequently travel at high speeds. 
This causes a discontinuous signal in time, 
the non-zero values suddenly becoming null 
as the probe vehicle or vehicles depart from that particular location.
Theoretically, this sort of movement can be loosely inferred by looking at the heading and speed information
and knowing a priori the structure of the roads.
See figure \label{fig:displacement} for an illustration of this case.
The following models are attempts at building models that incorporate this observation.

\begin{figure}
  \centering
  \includegraphics[width=\textwidth]{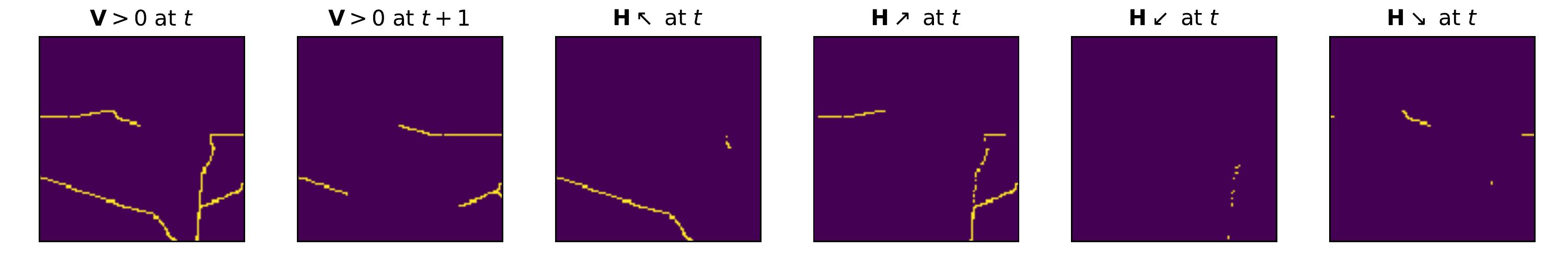}
  \caption{%
    Volume ($\mathbf{V}$) and heading ($\mathbf{H}$) information for the four directions
    (north-west $\nwarrow$, north-east $\nearrow$, south-west $\swarrow$, south-east $\searrow$)
    at two consecutive timestamps ($t$ and $t + 1$) on
    a 100 $\times$ 100-pixel crop from Berlin.
    Notice that the location of the probe vehcile in the central, horizontal road at frame $t + 1$
    can be deduced from the previous frame (time $t$) by using the corresponding heading information,
    which implies that the vehicle is moving towards the east.}
  \label{fig:displacement}
\end{figure}

\subsection{Adapting the filters}
\label{subsec:adapated-filters}
 
Arguably the most straightforward way of incorporating spatial information into the model is through 2d convolutions with a kernel size greater than one.
However, we observed that when the model has a large receptive field the performance degrades, as the model starts predicting more and more zeros.
We believe that there are a couple of reasons for this behaviour:
\ia the assumption made by the convolution layer of translation invariance is violated, since at different locations we have different road configurations;
\ib given that at any time most of the pixels in a frame are zero, the model relies on the simplest solution---predict always zeros.
In order to alleviate these issues we have tried adapting the filters based on their location.

The first attempt was to use \textit{local filters}.
This type of layer shares similarities with convolutional and fully connected layers:
it aggregates local information from neighbouring pixels (similar to convolutions) and it uses different weights for each output location (similar to fully connected layers).
The local filters can also be regarded as a parametric version of a spatio-temporal nearest neighbour classifier,
which is a popular solution for traffic forecasting \cite{cai_spatiotemporal_2016}.
The implementation of this layer was problematic, since such an operation is not very memory efficient.
We were, however, able to use it with a small number of intermediate channels.
Unfortunately, the network tended to overfit and it did not generalize better than the temporal regression model.
The corresponding code can be found in the \codemodel{Marcus} and \codemodel{Conv2dLocal} classes from \texttt{models/nn.py}.

The second attempt relies on \textit{dynamic filters} and tries to mitigate the overfitting problem by constraining the weights of the local filters to depend on the input.
The idea is to use a convolutional network model to dynamically generate the parameters for our model based for each prediction;
this method is similar to the dynamic filters networks proposed by De Brabandere \etal \cite{de_brabandere_dynamic_2016}.
The input of the parameter generator network is composed of the input to the predictor model (\ie, volume, speed and heading)
to which we add position information (\ie, $(x,y)$ coordinates).
The corresponding code can be found in the \codemodel{Nero2} class from \texttt{models/nn.py}.

\subsection{Learning the displacement}
\label{subsec:learning-displacement}

Based on the initial observation,
we model the location of the probe vehicles using a mask and
we predict their future location by learning a displacement of the mask.
The mask $\mathbf{M}$ is a frame with binary values:
it is one where there is at least some volume ($\mathbf{V} \ge 1$) 
and zero otherwise.
The displacement is predicted with a network which takes as input information such as heading and speed
and outputs two values for each location of the frame $(\Delta_x, \Delta_y)$.
The mask is warped using the displacement field and is then used to modulate the predictions $\mathbf{Y}$ of a usual network (\eg, a temporal regression network).
The final prediction is obtained by multiplying element-wise the two predictions,
$\mathbf{Y}' = \mathbf{M} \odot \mathbf{Y}$.
This method is similar to the spatial transformer network \cite{jaderberg_spatial_2016} and the implementation is based on Pytorch's \texttt{grid\_sample} function.
We use three losses on each of the three quantities:
\begin{itemize}
    \item on the final predictions $\mathbf{Y}'$ we use MSE;
    \item on the predicted mask $\mathbf{M}$ we use cross-entropy;
    \item on the predicted values $\mathbf{Y}$ we use MSE, but computed only at those location where there is some volume.
\end{itemize}

Associated code can be found on the \texttt{doneata/44/mask-displacements} branch, in the class \codemodel{Pomponia}.
As stated before, we haven't seen improvements using this method, and we believe the problem of learning the displacement from scratch is inherently difficult.

\subsection{Other observations}

In this subsection, we briefly state a few more empirical findings.

First, the following changes to the model did not seem to affect the performance in a significant way:
\begin{itemize}
    \item training independently on channels;
    \item predicting one frame, followed by a running average to predict the next;
    \item predicting discrete values for the heading channel, followed by a weighted average.
\end{itemize}

Second, we have also tried to augment the input with seasonal information (data from the previous days at the same time).
However, the improvements were marginal, while complicating the pipeline.
In the end, we decided not to resort to this sort of information.

\section{Conclusions}
\label{sec:conclusions}

We have presented our submission to the Traffic4cast competition.
Our model consists of a convolutional network augmented with biases that incorporate global spatial and temporal information.
The architecture is straightforward and has the advantage of being fast to train while obtaining good performance.
Notably, we observed that more intricate models fail to improve over the main system,
highlighting the challenge of finding suitable architectures that incorporate knowledge of the system's dynamics.
\bibliographystyle{unsrt}
\bibliography{ref}

\begin{thebibliography}{10}

\bibitem{cai_spatiotemporal_2016}
Pinlong Cai, Yunpeng Wang, Guangquan Lu, Peng Chen, Chuan Ding, and Jianping
  Sun.
\newblock A spatiotemporal correlative k-nearest neighbor model for short-term
  traffic multistep forecasting.
\newblock {\em Transportation Research Part C: Emerging Technologies},
  62:21--34, January 2016.

\bibitem{cheng_deeptransport:_2018}
Xingyi Cheng, Ruiqing Zhang, Jie Zhou, and Wei Xu.
\newblock {DeepTransport}: {Learning} spatial-temporal dependency for traffic
  condition forecasting.
\newblock In {\em 2018 {International} {Joint} {Conference} on {Neural}
  {Networks} ({IJCNN})}, pages 1--8. IEEE, 2018.

\bibitem{li_diffusion_2018}
Yaguang Li, Rose Yu, Cyrus Shahabi, and Yan Liu.
\newblock Diffusion {Convolutional} {Recurrent} {Neural} {Network}:
  {Data}-{Driven} {Traffic} {Forecasting}.
\newblock In {\em International {Conferences} on {Learning} {Representations}},
  February 2018.

\bibitem{xue_visual_2016}
Tianfan Xue, Jiajun Wu, Katherine Bouman, and Bill Freeman.
\newblock Visual dynamics: {Probabilistic} future frame synthesis via cross
  convolutional networks.
\newblock In {\em Advances in Neural Information Processing Systems}, pages
  91--99, 2016.

\bibitem{zhou_view_2016}
Tinghui Zhou, Shubham Tulsiani, Weilun Sun, Jitendra Malik, and Alexei~A Efros.
\newblock View synthesis by appearance flow.
\newblock In {\em European conference on computer vision}, pages 286--301.
  Springer, 2016.

\bibitem{vondrick_generating_2017}
Carl Vondrick and Antonio Torralba.
\newblock Generating the future with adversarial transformers.
\newblock In {\em Proceedings of the {IEEE} {Conference} on {Computer} {Vision}
  and {Pattern} {Recognition}}, pages 1020--1028, 2017.

\bibitem{yu_spatiotemporal_2017}
Haiyang Yu, Zhihai Wu, Shuqin Wang, Yunpeng Wang, and Xiaolei Ma.
\newblock Spatiotemporal recurrent convolutional networks for traffic
  prediction in transportation networks.
\newblock {\em Sensors}, 17(7):1501, 2017.

\bibitem{clevert_fast_2016}
Djork-Arné Clevert, Thomas Unterthiner, and Sepp Hochreiter.
\newblock Fast and {Accurate} {Deep} {Network} {Learning} by {Exponential}
  {Linear} {Units} ({ELUs}).
\newblock {\em arXiv:1511.07289 [cs]}, February 2016.
\newblock arXiv: 1511.07289.

\bibitem{li_hyperband:_2018}
Lisha Li, Kevin Jamieson, Giulia DeSalvo, Afshin Rostamizadeh, and Ameet
  Talwalkar.
\newblock Hyperband: {A} {Novel} {Bandit}-{Based} {Approach} to
  {Hyperparameter} {Optimization}.
\newblock {\em Journal of Machine Learning Research}, 18(185):1--52, 2018.

\bibitem{de_brabandere_dynamic_2016}
Bert De~Brabandere, Xu~Jia, Tinne Tuytelaars, and Luc Van~Gool.
\newblock Dynamic filter networks.
\newblock In {\em Advances in Neural Information Processing Systems}, pages
  667--675. Curran Associates Inc., 2016.

\bibitem{jaderberg_spatial_2016}
Max Jaderberg, Karen Simonyan, Andrew Zisserman, and Koray Kavukcuoglu.
\newblock Spatial {Transformer} {Networks}.
\newblock {\em arXiv:1506.02025 [cs]}, February 2016.
\newblock arXiv: 1506.02025.

\end{thebibliography}

\end{document}